\documentclass[10pt, a4paper]{article}
\usepackage[final]{lrec2026}

\usepackage{booktabs}
\usepackage{multirow}
\usepackage{xspace}
\usepackage{amsmath}
\usepackage{amssymb}
\usepackage{graphicx}
\usepackage{tikz}
\usepackage{pgfplots}
\pgfplotsset{compat=1.18}
\usepgfplotslibrary{fillbetween}
\usepackage{tcolorbox}
\usepackage{stfloats}
\usetikzlibrary{shapes.geometric, arrows.meta, positioning, fit, backgrounds, calc}

\newcommand{\method}{OntoBook\xspace}

\title{OntoBook: Ontology-Grounded Synthetic Textbooks for Medical Encoder Pretraining}

\name{Rian Touchent, Éric de La Clergerie}

\address{Inria, Sorbonne Université \\
         48 rue Barrault 75013 Paris, 21 rue de l'école de médecine 75006 Paris \\
         \{rian.touchent,eric.de\_la\_clergerie\}@inria.fr\\}

\abstract{
We present \method, a method that converts medical ontology structure into pretraining signal for encoder language models. Our approach has three stages: random walks through ontology graphs capture hierarchical and causal relations between medical codes, a large language model reformulates these walks into fluent textbook-style prose, and the resulting text is used to train ModernCamemBERT, a 149M-parameter French encoder, with two objectives on the same data: masked language modeling and relation prediction between code pairs. On three French medical coding benchmarks (FRACCO, Cantemist-FR, Distemist-FR), \method achieves significant improvements over MLM-only pretraining, with +2.5 micro-F1 on FRACCO and +8.0 micro-F1 on Distemist. We find that alignment between objectives is necessary: misaligned training, where each task uses different data, causes a 30-point degradation. We release 1.3 million LLM-reformulated medical textbooks across three French ontologies (CIM-10, CCAM, ATC) and pretrained model checkpoints.
\\ \newline 

}

\begin{document}

\maketitleabstract

\begin{figure*}[b]
\centering
\begin{tikzpicture}[
  lbl/.style={font=\scriptsize\sffamily, text=black!50, inner sep=1pt},
  stage/.style={font=\scriptsize\sffamily\bfseries, text=#1!70},
  frame/.style={rounded corners=5pt, draw=black!15, minimum height=2.6cm},
]

  \node[frame, minimum width=2.5cm]  (f1) at (1.25, 1.0) {};
  \node[frame, minimum width=5.0cm]  (f2) at (6.5,  1.0) {};
  \node[frame, minimum width=4.5cm]  (f3) at (12.75,1.0) {};

  \node[stage=blue]   at (f1.north) [above=2pt] {Ontology};
  \node[stage=orange] at (f2.north) [above=2pt] {Textbook {\scriptsize(1.3B tokens)}};
  \node[stage=red]    at (f3.north) [above=2pt] {Aligned training};

  \node[rectangle, rounded corners=3pt, draw=blue!50, fill=blue!8, inner sep=2.5pt,
        font=\scriptsize\ttfamily, minimum width=0.7cm] (e11) at ([xshift=-8mm,yshift=5mm]f1.center) {E11};
  \node[rectangle, rounded corners=3pt, draw=blue!50, fill=blue!8, inner sep=2.5pt,
        font=\scriptsize\ttfamily, minimum width=0.7cm] (n08) at ([xshift=7mm,yshift=5mm]f1.center) {N08};
  \node[rectangle, rounded corners=3pt, draw=blue!50, fill=blue!8, inner sep=2.5pt,
        font=\scriptsize\ttfamily, minimum width=0.7cm] (n083) at ([xshift=7mm,yshift=-5mm]f1.center) {N08.3};
  \draw[-{Stealth[length=1.5mm]}, blue!40, thick]
    (e11) -- node[above=-1pt, font=\tiny\sffamily, text=black!50] {\textsc{causes}} (n08);
  \draw[-{Stealth[length=1.5mm]}, blue!40, thick]
    (n08) -- node[left=-1pt, font=\tiny\sffamily, text=black!50] {\textsc{parent}} (n083);

  \node[draw=orange!40, fill=orange!4, rounded corners=3pt,
        text width=4.4cm, font=\scriptsize, inner sep=4pt, align=justify] (para) at (f2.center)
    {Le diab\`ete (\textcolor{blue!70}{\textbf{E11}}) peut entra\^iner
     une glom\'erulopathie (\textcolor{blue!70}{\textbf{N08}}),
     notamment la forme diab\'etique (\textcolor{blue!70}{\textbf{N08.3}})\ldots};

  \node[draw=red!30, fill=red!4, rounded corners=2pt,
        text width=3.6cm, font=\scriptsize, inner sep=4pt] (mlmex) at ([yshift=5.2mm]f3.center)
    {$\mathcal{L}_{\text{MLM}}$:\; Le \colorbox{red!15}{[MASK]} (E11) peut
     \colorbox{red!15}{[MASK]} une n\'ephropathie\ldots};
  \node[draw=red!30, fill=red!4, rounded corners=2pt,
        text width=3.6cm, font=\scriptsize, inner sep=4pt] (relex) at ([yshift=-6.8mm]f3.center)
    {$\mathcal{L}_{\text{rel}}$:\qquad E11 $\xrightarrow{\;\colorbox{red!15}{\textbf{?}}\;}$ N08};

  \draw[-{Stealth[length=2.5mm]}, line width=1.2pt, draw=black!35]
    (f1.east) -- node[above, font=\tiny\sffamily, text=black!50] {Qwen3-235B} (f2.west);
  \draw[-{Stealth[length=2mm]}, line width=1pt, draw=black!35]
    (f2.east) -- node[above, font=\tiny\sffamily, text=black!50] {Multi-task} (f3.west);

\end{tikzpicture}
\caption{\method pipeline. An ontology subgraph is converted into a random walk, then reformulated into textbook prose by Qwen3-235B. The encoder trains on this text with two aligned objectives: masked language modeling and relation prediction between code pairs.}
\label{fig:pipeline}
\end{figure*}
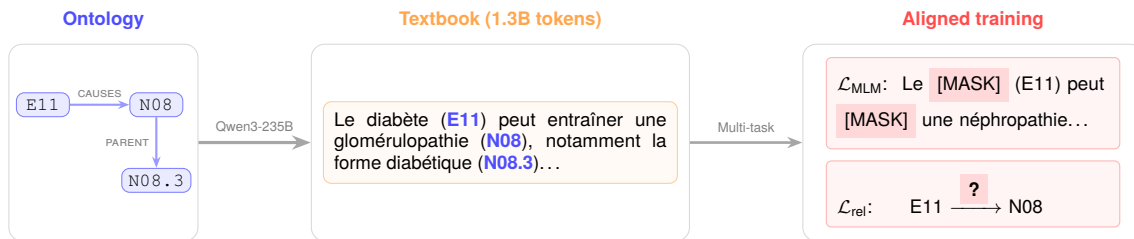

\section{Introduction}
\label{sec:intro}

Medical coding is the task of assigning standardized codes from medical ontologies to clinical documents. It is central to hospital billing, epidemiological surveillance, and clinical research. In France, over 30 million hospital stays per year require accurate coding using the CIM-10 classification (the French adaptation of ICD-10), CCAM procedure codes, and ATC medication codes. This task requires understanding not only medical vocabulary but also the relational structure of medical ontologies: hierarchical links between codes, causal associations, differential diagnoses, and exclusion rules. These relationships are explicitly encoded in ontology graphs but are absent from clinical text corpora.

Pretrained biomedical language models such as BioBERT \citep{lee2020biobert}, PubMedBERT \citep{gu2021pubmedbert}, and CamemBERT-bio \citep{touchent2023camembertbio} learn medical vocabulary from large text corpora. However, they do not capture the relational structure of medical ontologies, since this structure is not expressed in running text. Knowledge graph approaches such as RDF2Vec \citep{ristoski2016rdf2vec} and Snomed2Vec \citep{agarwal2019snomed2vec} encode ontology structure through random walks, but produce static, non-contextualized embeddings that cannot benefit from transformer pretraining. Knowledge-enhanced transformers such as DRAGON \citep{yasunaga2022dragon} and KEPLER \citep{wang2021kepler} integrate knowledge graph structure into language model pretraining, but they rely on existing text-graph alignment rather than generating new training data from ontology structure alone.

We propose \method, a pretraining method that converts medical ontology structure into training signal for encoder language models. Our approach proceeds in three stages. First, we generate random walks through ontology graphs to produce sequences that capture hierarchical, causal, and differential relationships between medical codes. Second, we reformulate these walks into fluent medical prose using a large language model, creating synthetic textbooks grounded in ontology structure. Third, we train an encoder with two objectives on the same textbook data: masked language modeling and relation prediction between code pairs. We refer to this co-training on shared data as \emph{alignment}, and our experiments show that it is a necessary condition for the method to work.

We evaluate \method on three French medical coding benchmarks and present ablation studies on the contribution of each component. We release 1.3 million LLM-reformulated medical textbooks across three French ontologies (CIM-10, CCAM, ATC) and all pretrained model checkpoints.

\section{Related Work}
\label{sec:related}

This section reviews three research directions: biomedical language model pretraining, knowledge-enhanced language models, and automatic medical coding.

\subsection{Biomedical Pretrained Language Models}

Domain-specific pretraining has proven essential for biomedical NLP. BioBERT \citep{lee2020biobert} further pretrained BERT on PubMed abstracts and PMC full-text articles, improving named entity recognition and relation extraction. PubMedBERT \citep{gu2021pubmedbert} demonstrated that pretraining from scratch on in-domain text outperforms mixed-domain initialization. ClinicalBERT \citep{huang2019clinicalbert} targeted clinical notes from MIMIC-III. For French, CamemBERT-bio \citep{touchent2023camembertbio} adapted CamemBERT \citep{martin2020camembert} to biomedical text, while DrBERT \citep{labrak2023drbert} was trained on French biomedical and clinical text from the NACHOS corpus. These models learn from flat text corpora and do not capture the hierarchical structure of medical ontologies.

\subsection{Knowledge-Enhanced Language Models}

Random walks on knowledge graphs generate sequences capturing relational structure. RDF2Vec \citep{ristoski2016rdf2vec} adapted the walk-then-embed paradigm to RDF graphs. In the biomedical domain, Snomed2Vec \citep{agarwal2019snomed2vec} applied random walk and Poincar\'e embeddings to SNOMED-CT for clinical prediction tasks. OWL2Vec* \citep{chen2021owl2vec} combined random walk sequences with lexical and logical features from OWL ontologies. These methods produce static embeddings (Word2Vec-style) and cannot be used for transformer pretraining.

Several approaches inject knowledge graph structure into transformers. ERNIE \citep{zhang2019ernie} aligns entity embeddings with text representations. K-BERT \citep{liu2020kbert} injects knowledge triples into the input sequence using soft-position indices and a visibility mask. For multi-task pretraining, KEPLER \citep{wang2021kepler} combines MLM with TransE-style knowledge embedding on Wikidata. DRAGON \citep{yasunaga2022dragon} jointly trains MLM and knowledge graph link prediction, with a biomedical variant on UMLS achieving +3\% on MedQA. CODER \citep{yuan2021coder} uses UMLS relation triplets for contrastive pretraining, and SAPBERT \citep{liu2021sapbert} uses contrastive learning on UMLS synonyms for biomedical entity linking. Closest to the present work, BioOntoBERT \citep{shashikumar2023bioontobert} generates sentences from biomedical ontologies using Onto2Sen templates and pretrains BERT with MLM on the resulting corpus. However, the template-based generation produces formulaic sentences (e.g., ``X is a type of Y'') rather than fluent prose, and the method uses only MLM without multi-task relation prediction.

Recent work has explored synthetic data generation from knowledge sources. The Phi series \citep{gunasekar2023phi} demonstrated that LLM-generated textbook-quality data enables strong small-model performance. EntiGraph \citep{yang2025entigraph} generates entity-centric synthetic data from knowledge sources but targets decoder language models rather than encoder pretraining.

\subsection{Automatic Medical Coding}

Medical coding assigns ICD, procedure, or medication codes to clinical text. CAML \citep{mullenbach2018explainable} introduced label-wise attention for explainable predictions. PLM-ICD \citep{huang2022plmicd} showed that fine-tuning pretrained encoders with label attention directly improves coding performance. \citet{kirchler2026grasp} recently demonstrated that LLM embeddings improve transferability of EHR-based predictions across countries and coding systems, highlighting the importance of encoding medical knowledge structure for cross-system generalization. These methods improve the classification head or the input representations but rely on generic encoders pretrained on flat text corpora.

Table~\ref{tab:related_comparison} summarizes how \method combines components not jointly present in prior work.

\begin{table}[t]
\centering
\small
\begin{tabular}{lccc}
\toprule
\textbf{Method} & \textbf{Walks} & \textbf{LLM} & \textbf{MLM+Rel} \\
\midrule
Snomed2Vec & \checkmark & & \\
BioOntoBERT & \checkmark & & \\
DRAGON & & & \checkmark \\
KEPLER & & & \checkmark \\
Phi-1 & & \checkmark & \\
\midrule
\textbf{\method} & \checkmark & \checkmark & \checkmark \\
\bottomrule
\end{tabular}
\caption{Comparison with related approaches. \method uniquely combines all three components for biomedical encoder pretraining.}
\label{tab:related_comparison}
\end{table}

\section{Method}
\label{sec:method}

\method converts medical ontology structure into pretraining signal through three stages: (1) generating random walks through ontology graphs, (2) reformulating these walks into fluent textbook prose using a large language model, and (3) training an encoder with multi-task learning combining masked language modeling and relation prediction. We first describe the ontologies used, then detail each stage. Figure~\ref{fig:pipeline} illustrates the full pipeline.

\subsection{Medical Ontologies}
\label{sec:ontologies}

We use three French medical ontologies published by the Agence du Num\'erique en Sant\'e (ANS) through its terminology server (SMT) in RDF/OWL format. CIM-10 FR PMSI is the French adaptation of the WHO's ICD-10, enriched by the Agence Technique de l'Information sur l'Hospitalisation (ATIH) for the French hospital information system (Programme de M\'edicalisation des Syst\`emes d'Information, PMSI). It contains 19,161 diagnostic codes organized in a 5-level hierarchy (chapters, blocks, categories, subcategories), with textual annotations including 23,282 synonyms, 8,171 inclusion notes, and 381 definitions. CIM-10 is the only ontology with semantic edges beyond the hierarchy: 1,317 causal edges (e.g., E11 type 2 diabetes \textit{causes} N08.3 diabetic glomerulopathy), 5,739 exclusion edges, and 341 manifestation edges, yielding a mean node degree of 2.77 compared to 2.00 for the purely hierarchical ontologies.

CCAM (Classification Commune des Actes M\'edicaux) is the French procedure classification maintained by the Caisse Nationale d'Assurance Maladie (CNAM), containing 38,191 procedure codes in a 4-level hierarchy with 8,991 synonyms and 1,188 disjointness constraints between mutually exclusive procedures. ATC (Anatomical Therapeutic Chemical) is the WHO medication classification with 6,950 substance codes in a strict 5-level hierarchy from anatomical group to chemical substance, with no semantic edges, no synonyms, and no definitions beyond labels. This contrast in relational richness directly shapes the walk generation strategy: CIM-10 walks can follow causal and differential diagnosis paths, whereas CCAM and ATC walks are limited to hierarchical and sibling traversals. Table~\ref{tab:ontology_stats} summarizes the structural properties of the three ontologies.

\begin{table}[t]
\centering
\small
\begin{tabular}{lrrr}
\toprule
& \textbf{CIM-10} & \textbf{CCAM} & \textbf{ATC} \\
\midrule
\multicolumn{4}{l}{\textit{Graph structure}} \\
Codes & 19,161 & 38,191 & 6,950 \\
Hierarchy depth & 5 & 4 & 5 \\
Hierarchical edges & 19,139 & 38,191 & 6,950 \\
Causal edges & 1,317 & --- & --- \\
Manifestation edges & 341 & --- & --- \\
Exclusion edges & 5,739 & --- & --- \\
Disjoint edges & --- & 1,188 & --- \\
Mean degree & 2.77 & 2.06 & 2.00 \\
\midrule
\multicolumn{4}{l}{\textit{Textual attributes}} \\
Synonyms & 23,282 & 8,991 & 0 \\
Inclusion notes & 8,171 & --- & --- \\
Definitions & 381 & 151 & 0 \\
\bottomrule
\end{tabular}
\caption{Structural properties of the three French medical ontologies used for walk generation. CIM-10 is the only ontology with semantic edges beyond the hierarchy.}
\label{tab:ontology_stats}
\end{table}

\subsection{Ontology Walk Generation}
\label{sec:walks}

We generate random walks through ontology graphs to capture relational structure in a sequential format suitable for language model pretraining. We sample walks starting from each code in the ontology. At each step, the next node is selected via weighted sampling where edge type weights depend on the walk type. This biased sampling is necessary because semantic edges are rare (Table~\ref{tab:ontology_stats}), and a uniform walk would rarely traverse them. Each walk records the traversed codes along with their labels and the relation types connecting them. Walk length is sampled uniformly between 8 and 20 steps. We track visited nodes to avoid cycles, with a fallback that allows revisiting nodes when all neighbors have been explored.

For CIM-10, we define five walk types, each emphasizing different aspects of medical knowledge: \textit{Etiologie} (causal chains), \textit{Diagnostic Diff\'erentiel} (codes to distinguish clinically), \textit{Codage Double} (mandatory code pairs linking etiology to manifestation), \textit{Syndrome} (multi-system manifestations), and \textit{Cross-Chapter} (connections between different ICD chapters, e.g., diabetes E11 to its renal complication N08). We generate 402,328 walks for CIM-10 (average 4,830 characters), 762,958 for CCAM, and 138,980 for ATC, totaling 1,304,266 walks.

\subsection{Textbook Generation}
\label{sec:textbook}

Raw walks contain structured markers (e.g., ``\texttt{>> \`A distinguer de:}'') that are useful for parsing but not natural for language model pretraining. We use a large language model to reformulate walks into fluent medical prose resembling textbook paragraphs.

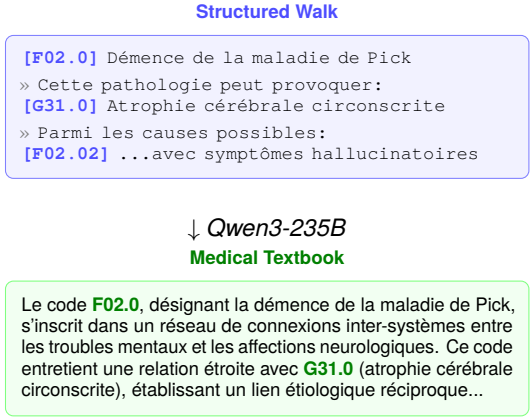
\begin{figure}[t]
\centering
\begin{tikzpicture}[node distance=0.3cm]
  \node[draw=blue!50, fill=blue!5, rounded corners=3pt,
        text width=6.5cm, align=left, inner sep=6pt,
        font=\scriptsize\ttfamily] (walk) {
    \textcolor{blue!70}{\textbf{[F02.0]}} D\'emence de la maladie de Pick\\[2pt]
    \textcolor{gray}{>>} Cette pathologie peut provoquer:\\
    \textcolor{blue!70}{\textbf{[G31.0]}} Atrophie c\'er\'ebrale circonscrite\\[2pt]
    \textcolor{gray}{>>} Parmi les causes possibles:\\
    \textcolor{blue!70}{\textbf{[F02.02]}} ...avec sympt\^omes hallucinatoires
  };
  \node[above=0.1cm of walk, font=\scriptsize\sffamily\bfseries, text=blue!70] {Structured Walk};
  \node[below=0.4cm of walk, font=\small] (arrow) {$\downarrow$ \textit{Qwen3-235B}};
  \node[draw=green!50, fill=green!5, rounded corners=3pt,
        text width=6.5cm, align=justify, inner sep=6pt,
        font=\scriptsize, below=0.4cm of arrow] (textbook) {
    Le code \textcolor{green!50!black}{\textbf{F02.0}}, d\'esignant la d\'emence de la maladie de Pick,
    s'inscrit dans un r\'eseau de connexions inter-syst\`emes entre les troubles mentaux et les
    affections neurologiques. Ce code entretient une relation \'etroite avec
    \textcolor{green!50!black}{\textbf{G31.0}} (atrophie c\'er\'ebrale circonscrite), \'etablissant
    un lien \'etiologique r\'eciproque...
  };
  \node[above=0.1cm of textbook, font=\scriptsize\sffamily\bfseries, text=green!50!black] {Medical Textbook};
\end{tikzpicture}
\caption{Walk-to-textbook transformation. The structured walk is reformulated into fluent medical prose while preserving all codes and relations.}
\label{fig:textbook_example}
\end{figure}

We prompt Qwen3-235B-A22B-Instruct \citep{yang2025qwen3} with FP8 quantization to transform each walk into a coherent paragraph. The prompt instructs the model to: (1) preserve all medical codes and their relationships exactly as stated, (2) use natural medical language without lists or bullet points, (3) not add any information beyond what appears in the walk. This ensures the textbook content remains grounded in the ontology structure. We apply two filters: we discard reformulations shorter than 50 characters or that fail to mention the source codes. The reformulation runs on 4 NVIDIA H100 GPUs using vLLM \citep{kwon2023vllm} with guided decoding to ensure valid JSON output, taking approximately 20 hours for all 1.3 million walks. The resulting textbooks contain approximately 1.3 billion tokens across all three ontologies. Figure~\ref{fig:textbook_example} shows an example transformation.

\begin{figure}[t]
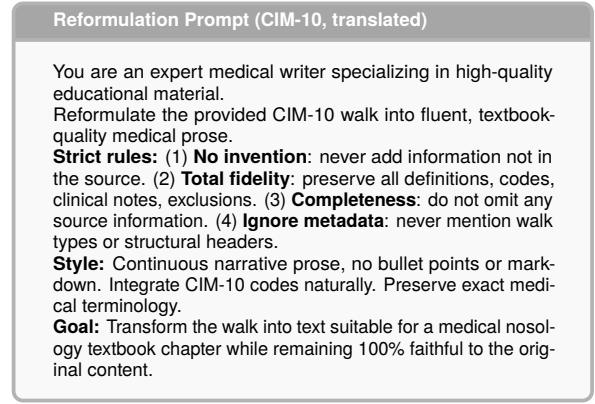

\begin{tcolorbox}[colback=gray!5, colframe=gray!60, fontupper=\scriptsize, title={\scriptsize\sffamily\bfseries Reformulation Prompt (CIM-10, translated)}, coltitle=white, colbacktitle=gray!70]
You are an expert medical writer specializing in high-quality educational material.

Reformulate the provided CIM-10 walk into fluent, textbook-quality medical prose.

\textbf{Strict rules:}
(1) \textbf{No invention}: never add information not in the source.
(2) \textbf{Total fidelity}: preserve all definitions, codes, clinical notes, exclusions.
(3) \textbf{Completeness}: do not omit any source information.
(4) \textbf{Ignore metadata}: never mention walk types or structural headers.

\textbf{Style:} Continuous narrative prose, no bullet points or markdown. Integrate CIM-10 codes naturally. Preserve exact medical terminology.

\textbf{Goal:} Transform the walk into text suitable for a medical nosology textbook chapter while remaining 100\% faithful to the original content.
\end{tcolorbox}
\caption{System prompt for LLM-based walk reformulation. Temperature is set to 0 for deterministic, faithful output.}
\label{fig:prompt}
\end{figure}

Figure~\ref{fig:prompt} shows the system prompt used for CIM-10 walks (translated from French for readability; prompts for CCAM and ATC follow the same structure with ontology-specific terminology).

\subsection{\method Training}
\label{sec:training}

We train a ModernCamemBERT-base encoder \citep{music2025moderncamembert}, initialized from a French checkpoint trained on 1 trillion tokens.

The model optimizes two objectives jointly. The first, $\mathcal{L}_{\text{MLM}}$, follows standard masked language modeling \citep{devlin2019bert}: we mask 15\% of tokens from textbook paragraphs and train to predict them. The second, $\mathcal{L}_{\text{rel}}$, classifies the relationship between pairs of medical codes. Our main model uses CIM-10; we explore transfer from CCAM and ATC in Section~\ref{sec:multi_onto}. We extract 282,907 code pairs from the CIM-10 ontology across 6 relation types: \textsc{Parent}, \textsc{Child}, \textsc{Sibling}, \textsc{Differential Diagnosis}, \textsc{Causes}, and \textsc{Caused\_By}. For each pair, we retrieve the corresponding textbook descriptions generated in Section~\ref{sec:textbook}. The input format is "$[\texttt{CLS}]~\text{text}_A~[\texttt{SEP}]~\text{text}_B~[\texttt{SEP}]$", where $\text{text}_A$ and $\text{text}_B$ are the textbook descriptions of the two codes, and a separate linear classification head on the $[\texttt{CLS}]$ token predicts the relation type, trained with cross-entropy loss over the 6 classes.

The combined loss is:
\begin{equation}
\mathcal{L} = \mathcal{L}_{\text{MLM}} + \lambda \mathcal{L}_{\text{rel}}
\end{equation}
where $\lambda = 1.0$. We train for 4 epochs with learning rate $2 \times 10^{-5}$, batch size 128, and AdamW optimizer with weight decay 0.01 and 1,000 warmup steps. Training takes approximately 4 hours per epoch on one NVIDIA H100 GPU. The key design choice is that both objectives operate on the \emph{same textbook data}. We call this property \emph{alignment}, and we show in Section~\ref{sec:ablations} that it is essential for performance.

\section{Experiments}
\label{sec:experiments}

\subsection{Experimental Setup}
\label{sec:setup}

\paragraph{Evaluation Benchmarks.}
We evaluate on the standard French medical coding evaluation suite:
\begin{itemize}
\item \textbf{FRACCO} \citep{pignat2024fracco}: a native French oncology corpus with 1,301 clinical cases annotated with ICD-O-3.1 codes, framed as multi-label classification over the top 100 codes.
\item \textbf{Cantemist-FR} \citep{zaghir2023frasimed}: the French adaptation of the Spanish tumor coding task \citep{miranda2020cantemist}, with 2,051 clinical documents.
\item \textbf{Distemist-FR} \citep{zaghir2023frasimed}: the French adaptation of the disease mention coding task \citep{miranda2022distemist}.
\end{itemize}
Cantemist-FR and Distemist-FR are cross-lingual projections from Spanish. FRACCO is the only native French benchmark. All results are micro-F1 scores averaged over 3 random seeds.

\paragraph{Baselines.}
We compare against French biomedical language models: CamemBERT-bio \citep{touchent2023camembertbio}, DrBERT \citep{labrak2023drbert}, and ModernCamemBERT \citep{music2025moderncamembert} (which serves as our initialization checkpoint). We also report results for three ablations of our method: MLM-only (trained with $\mathcal{L}_{\text{MLM}}$ only on our textbooks), Rel-only (trained with $\mathcal{L}_{\text{rel}}$ only), and CodeInfill+MLM, which replaces relation prediction with a code infilling objective that masks medical code tokens (e.g., ``E11'', ``N08.3'') in textbook text and trains the model to predict them. We do not compare against knowledge-enhanced models such as DRAGON \citep{yasunaga2022dragon} or SAPBERT \citep{liu2021sapbert} as they are English encoders that cannot be applied to French text.

\paragraph{Fine-tuning.}
For downstream evaluation, each pretrained encoder receives a clinical document as input and predicts a set of medical codes (multi-label classification over the top 100 codes per benchmark). We fine-tune with a linear classification head for 10 epochs, learning rate $2 \times 10^{-5}$, and batch size 16. External baselines (CamemBERT-bio, DrBERT, ModernCamemBERT) are averaged over 9 seeds; our models over 3 seeds due to the large number of configurations evaluated.

\subsection{Main Results}
\label{sec:results}

Table~\ref{tab:main_results} compares \method against French biomedical baselines.

\begin{table}[t]
\centering
\resizebox{\columnwidth}{!}{%
\begin{tabular}{lcccc}
\toprule
\textbf{Model} & \textbf{FRACCO} & \textbf{Cant.} & \textbf{Dist.} & \textbf{Avg.} \\
\midrule
\multicolumn{5}{l}{\textit{External baselines}} \\
CamemBERT-bio & 20.2{\scriptsize$\pm$0.2} & 12.1{\scriptsize$\pm$0.4} & 9.0{\scriptsize$\pm$0.2} & 13.8 \\
DrBERT & 36.3{\scriptsize$\pm$0.7} & 37.7{\scriptsize$\pm$1.0} & 22.5{\scriptsize$\pm$0.7} & 32.1 \\
ModernCamemBERT & 56.4{\scriptsize$\pm$1.0} & 63.5{\scriptsize$\pm$1.3} & 23.4{\scriptsize$\pm$1.7} & 47.7 \\
\midrule
\multicolumn{5}{l}{\textit{Our models}} \\
CodeInfill+MLM & 56.5{\scriptsize$\pm$0.2} & 66.4{\scriptsize$\pm$1.9} & 20.0{\scriptsize$\pm$0.9} & 47.6 \\
MLM-only & 55.8{\scriptsize$\pm$0.4} & 66.0{\scriptsize$\pm$1.1} & 24.2{\scriptsize$\pm$5.8} & 48.7 \\
\textbf{\method} & \textbf{58.3}{\scriptsize$\pm$0.3} & \textbf{67.1}{\scriptsize$\pm$1.1} & \textbf{32.2}{\scriptsize$\pm$1.1} & \textbf{52.5} \\
\bottomrule
\end{tabular}%
}
\caption{Main results on French medical coding (micro-F1 \%). External baselines averaged over 9 seeds; our models over 3 seeds. Best results in bold.}
\label{tab:main_results}
\end{table}

\method outperforms all baselines on all three benchmarks. The improvement over MLM-only pretraining is significant on FRACCO (+2.5 micro-F1, $p<0.001$) and Distemist (+8.0 micro-F1, $p<0.001$), with consistent but not individually significant gains on Cantemist ($p=0.11$). The gain is largest on Distemist, suggesting that ontology-aware pretraining particularly helps with fine-grained disease coding. \method also reduces variance on Distemist from $\pm$5.8 (MLM-only) to $\pm$1.1.

\subsection{Ablation Study}
\label{sec:ablations}

Table~\ref{tab:ablations} ablates the key components of \method.

\begin{table}[t]
\centering
\resizebox{\columnwidth}{!}{%
\begin{tabular}{lccccc}
\toprule
\textbf{Configuration} & \textbf{FRACCO} & \textbf{Cant.} & \textbf{Dist.} & \textbf{Avg.} & \textbf{$\Delta$} \\
\midrule
\method (aligned) & 58.33 & 67.06 & 32.24 & 52.54 & --- \\
\quad $-$ $\mathcal{L}_{\text{rel}}$ (MLM-only) & 55.81 & 66.01 & 24.23 & 48.68 & $-$3.86 \\
\quad $-$ $\mathcal{L}_{\text{MLM}}$ (Rel-only) & 48.24 & 51.15 & 20.18 & 39.86 & $-$12.68 \\
\quad $-$ Alignment (misaligned) & 33.39 & 20.11 & 12.63 & 22.04 & $-$30.50 \\
\midrule
MLM-only (raw walks) & 55.51 & 66.00 & 22.76 & 48.09 & $-$4.45 \\
\bottomrule
\end{tabular}%
}
\caption{Ablation study (micro-F1 \%). Misalignment degrades all benchmarks, with the largest drop on Cantemist ($-$46.95). All variants use the same base model and training budget.}
\label{tab:ablations}
\end{table}

Alignment is the most critical factor. Training $\mathcal{L}_{\text{MLM}}$ and $\mathcal{L}_{\text{rel}}$ on different data causes catastrophic degradation across all benchmarks, with Cantemist dropping from 67.06 to 20.11 ($-$46.95). Relation prediction alone also fails ($-$12.68 average F1), confirming that $\mathcal{L}_{\text{rel}}$ cannot serve as a standalone pretraining objective. The last row trains MLM-only on raw structured walks, removing both reformulation and relation prediction. The raw walk model (48.09) is close to the textbook MLM-only model (48.68), indicating that reformulation alone contributes little to MLM pretraining ($+$0.59 F1). However, the gap to full \method is 4.45 points, reflecting the combined effect of adding both reformulation and relation prediction. This suggests that reformulation primarily benefits the relation prediction objective: on raw walks with structural markers (e.g., ``\texttt{>> \`A distinguer de:}''), the classifier can exploit formatting shortcuts rather than learning medical semantics, whereas fluent textbook prose forces the model to learn from content rather than format.

\subsection{Training Dynamics and Hyperparameter Sensitivity}
\label{sec:dynamics}

\begin{figure}[t]
\centering
\begin{tikzpicture}
\begin{axis}[
  width=0.48\columnwidth, height=3.5cm,
  xlabel={\scriptsize Epoch},
  ylabel={\scriptsize Avg F1},
  xmin=0.5, xmax=3.5,
  ymin=47, ymax=54,
  xtick={1,2,3},
  ytick={48,50,52},
  grid=major,
  grid style={gray!20},
  mark size=2pt,
  every axis label/.style={font=\scriptsize},
  every tick label/.style={font=\scriptsize},
  title={\scriptsize\textbf{(a) Training epochs}},
  title style={at={(0.5,1.05)}},
]
\addplot[blue, thick, mark=*] coordinates {(1,51.98) (2,52.54) (3,49.43)};
\addplot[dashed, gray, thick] coordinates {(0.5,48.68) (3.5,48.68)};
\node[font=\tiny, gray] at (axis cs:2.9,47.8) {MLM-only};
\end{axis}
\end{tikzpicture}%
\hfill%
\begin{tikzpicture}
\begin{axis}[
  width=0.48\columnwidth, height=3.5cm,
  xlabel={\scriptsize $\lambda$},
  ylabel={\scriptsize FRACCO F1},
  xmode=log,
  xmin=0.07, xmax=7,
  ymin=56, ymax=61,
  xtick={0.1,0.5,1,2,5},
  xticklabels={0.1,0.5,1,2,5},
  ytick={57,58,59,60},
  grid=major,
  grid style={gray!20},
  mark size=2pt,
  every axis label/.style={font=\scriptsize},
  every tick label/.style={font=\tiny},
  title={\scriptsize\textbf{(b) Loss weight $\lambda$}},
  title style={at={(0.5,1.05)}},
]
\addplot[red, thick, mark=square*] coordinates {(0.1,58.95) (0.5,57.72) (1.0,58.66) (2.0,58.33) (5.0,59.08)};
\addplot[red, thick, mark=none, name path=upper, draw=none] coordinates {(0.1,60.34) (0.5,58.78) (1.0,59.18) (2.0,58.99) (5.0,59.88)};
\addplot[red, thick, mark=none, name path=lower, draw=none] coordinates {(0.1,57.56) (0.5,56.66) (1.0,58.14) (2.0,57.67) (5.0,58.28)};
\addplot[red!20, opacity=0.5] fill between[of=upper and lower];
\end{axis}
\end{tikzpicture}
\caption{Training dynamics. (a) Average F1 peaks at epoch 2 then degrades. (b) Performance is stable across $\lambda$ values (shaded: $\pm$1 std).}
\label{fig:dynamics}
\end{figure}
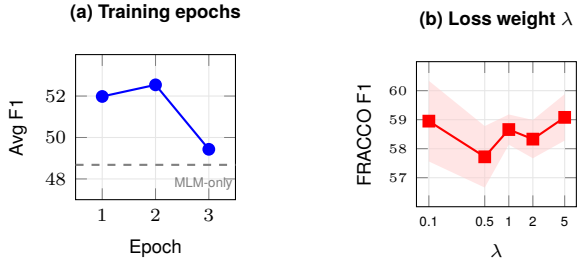

Figure~\ref{fig:dynamics} shows training dynamics and hyperparameter sensitivity. Epoch 1 (51.98) and epoch 2 (52.54) both outperform MLM-only pretraining (dashed line), but epoch 3 (49.43) degrades below it, possibly due to overfitting on the relation prediction task. All results in this paper use the epoch 2 checkpoint. The loss weight $\lambda$ is robust across $\{0.1, 0.5, 1.0, 2.0, 5.0\}$: all values achieve 57.7--59.1\% F1 on FRACCO. We use $\lambda=1.0$ as it achieves the lowest variance ($\pm$0.52\%). This robustness suggests that the exact balance between objectives matters less than ensuring both are present and aligned.

\subsection{Multi-Ontology Transfer}
\label{sec:multi_onto}

The main \method model uses CIM-10 walks (Table~\ref{tab:main_results}). To test whether other ontologies also provide useful pretraining signal, we train separate models on walks from each ontology individually and on all three combined.

\begin{table}[t]
\centering
\resizebox{\columnwidth}{!}{%
\begin{tabular}{lcccc}
\toprule
\textbf{Ontology source} & \textbf{FRACCO} & \textbf{Cant.} & \textbf{Dist.} & \textbf{Avg.} \\
\midrule
\textbf{CIM-10} (\method) & 58.33 & 67.06 & 32.24 & 52.54 \\
\midrule
All (CIM-10+CCAM+ATC) & 59.31 & \textbf{67.64} & 31.44 & 52.80 \\
CCAM only & 58.59 & 66.54 & 34.26 & \textbf{53.13} \\
ATC only & \textbf{59.66} & 63.03 & \textbf{36.05} & 52.91 \\
\bottomrule
\end{tabular}%
}
\caption{Effect of ontology source on downstream performance (micro-F1 \%). Each row trains a separate model using walks and relations from only that ontology. The main \method model uses CIM-10 only. All models use the same training procedure and base checkpoint.}
\label{tab:ontology_results}
\end{table}

All three individual ontologies improve over the MLM-only baseline, indicating that the pipeline generalizes across different medical knowledge structures. ATC (medications) yields the best Distemist score (+3.81 over CIM-10), despite Distemist being a disease coding task, suggesting that drug-disease relationships transfer well to disease understanding. CCAM (procedures) achieves the best average micro-F1 (53.13). Combining all three ontologies does not outperform the best individual ontologies. We hypothesize that this is because the relation prediction head must accommodate heterogeneous edge types across ontologies.

\subsection{Probing Analysis}
\label{sec:probing}

To understand how \method organizes code representations, we evaluate with three probing tasks on CIM-10 code embeddings (mean-pooled over code descriptions): chapter classification (26 classes), hierarchy depth prediction (4 levels), and distance correlation (Spearman $\rho$ between cosine distances and ontology tree distances). We compare against a code infilling variant (CodeInfill+MLM) that masks and predicts code tokens during pretraining.

\begin{table}[t]
\centering
\small
\begin{tabular}{lccc}
\toprule
\textbf{Model} & \textbf{Chapter} & \textbf{Depth} & \textbf{Dist. $\rho$} \\
\midrule
Base & 97.5\% & 99.3\% & 0.195 \\
MLM-only & 98.9\% & 99.4\% & 0.287 \\
CodeInfill+MLM & \textbf{99.0\%} & \textbf{99.8\%} & \textbf{0.397} \\
Rel-only & 98.0\% & 98.2\% & 0.203 \\
\method & 98.7\% & 99.7\% & 0.073 \\
\bottomrule
\end{tabular}
\caption{Probing analysis on CIM-10 code embeddings. CodeInfill+MLM achieves the best geometric organization but \method achieves the best downstream performance (Table~\ref{tab:main_results}).}
\label{tab:probing}
\end{table}

CodeInfill+MLM achieves the best probing metrics ($\rho=0.397$) but the worst downstream performance among our models (47.6 average micro-F1, Table~\ref{tab:main_results}), while \method has lower distance correlation ($\rho=0.073$) despite achieving the best downstream performance (52.5 micro-F1). This inverse relationship suggests that for encoder pretraining, representational flexibility is more valuable than strict geometric preservation of ontology structure.
\section{Discussion}
\label{sec:discussion}

The 30.50-point gap between aligned and misaligned training highlights the central role of objective alignment. When $\mathcal{L}_{\text{MLM}}$ trains on textbooks while $\mathcal{L}_{\text{rel}}$ trains on unrelated data, the model receives conflicting gradient signals: the MLM objective pushes representations toward textbook language patterns, while the relation objective pushes them toward a different data distribution. Aligned training avoids this conflict by ensuring both objectives reinforce the same semantic structure. Relation prediction guides attention toward ontologically meaningful patterns in the textbook text, while MLM ensures the model builds coherent representations of that text. This explains why relation prediction alone fails ($-$12.68 F1): it is a discriminative task that can exploit surface-level shortcuts, as demonstrated by the small gap between raw-walk and textbook MLM-only models (+0.59 F1). Combined with aligned MLM, however, it provides a complementary signal that improves downstream coding performance.

The cross-ontology transfer results suggest that our approach generalizes beyond CIM-10: training on CCAM or ATC alone also improves over the baseline despite targeting different medical domains. However, combining all ontologies does not outperform individual ones, possibly because the relation prediction head must accommodate heterogeneous edge types.

The probing analysis reveals a tension between geometric organization and downstream performance. CodeInfill+MLM achieves the best probing metrics ($\rho=0.397$) but the worst downstream scores among our models (47.6 micro-F1), while \method achieves the opposite ($\rho=0.073$, 52.5 micro-F1). This suggests that for downstream coding tasks, flexible representations adapt better than rigid geometric structure that must be partially unlearned during fine-tuning.

LLM reformulation plays a dual role. For MLM alone, the effect is modest ($+$0.59 F1), but reformulation is essential for relation prediction: on raw walks, structural markers provide formatting shortcuts that the classifier can exploit without learning medical semantics, whereas fluent prose forces the model to learn from content rather than format. The full gap between \method and MLM-only on raw walks is 4.45 points.

\paragraph{Limitations.}
All experiments use French medical coding. Extending to English ontologies such as ICD-11 and SNOMED-CT is left for future work. The reformulation step requires significant compute for LLM inference, and sensitivity to the choice of LLM has not been evaluated. Cantemist-FR and Distemist-FR are cross-lingual projections from Spanish \citep{zaghir2023frasimed}, which may introduce translation artifacts; FRACCO is the only native French benchmark. All experiments use ModernCamemBERT; generalization to other architectures remains untested.

\paragraph{Future Work.}
Future work will extend \method to English medical coding using UMLS and SNOMED-CT ontologies. A promising direction is cross-ontology walks that traverse edges between different ontologies (e.g., from a CIM-10 diagnosis to its ATC treatment to the CCAM procedure), generating richer walks that capture inter-ontology relationships in a single sequence.

\section{Conclusion}
\label{sec:conclusion}

We presented \method, a method that converts medical ontology structure into pretraining signal for encoder language models through random walks, LLM-based textbook reformulation, and multi-task learning combining masked language modeling and relation prediction. Our key finding is that alignment between objectives is essential: both tasks must train on the same data, as misalignment causes a 30-point degradation. On French medical coding benchmarks, \method improves over MLM-only pretraining by +2.5 micro-F1 on FRACCO and +8.0 on Distemist. We release 1.3 million LLM-reformulated medical textbooks across three French ontologies (CIM-10, CCAM, ATC) and pretrained model checkpoints.

\section*{Acknowledgements}
This work was granted access to the HPC resources of IDRIS under the allocation 2025-AD011014393R2 made by GENCI.

\section{Bibliographical References}\label{sec:reference}
\bibliographystyle{lrec2026-natbib}
\bibliography{references}

\end{document}